\documentclass[preprint,12pt]{elsarticle}

\usepackage{amssymb}
\usepackage{amsmath}
\usepackage[utf8]{inputenc} 
\usepackage[T1]{fontenc}    

\usepackage{url}            
\usepackage{booktabs}       
\usepackage{amsfonts}       
\usepackage{nicefrac}       
\usepackage{microtype}      
\usepackage{xcolor}         
\usepackage[normalem]{ulem}
\usepackage{natbib}
\usepackage{multirow}
\usepackage{graphicx}
\usepackage{subcaption}
\usepackage{CJKutf8}
\usepackage{etoolbox}
\usepackage{array}
\BeforeBeginEnvironment{tabular}{\fontsize{8pt}{10pt}\selectfont}
\newdefinition{definition}{Definition}
\newdefinition{rmk}{Remark}
\usepackage[ruled,vlined]{algorithm2e}

\usepackage{lineno}
\usepackage{hyperref}       
\usepackage{float}

\begin{document}

\begin{frontmatter}



\title{Enhancing Rare Codes via Probability-Biased Directed Graph Attention for Long-Tail ICD Coding}

\author[2]{Tianlei Chen}
\ead{chentianlei@ruc.edu.cn}
\author[2]{Yuxiao Chen}
\ead{cyx2023201820@ruc.edu.cn}
\author[1,2]{Yang Li}
\ead{yang.li@ruc.edu.cn}
\author[1,2]{Feifei Wang\corref{cor}}
\ead{feifei.wang@ruc.edu.cn}

\cortext[cor]{Corresponding author}

\affiliation[1]{organization={Center for Applied Statistics, Renmin University of China},
city={Beijing},
postcode={100872},
country={China}}
\affiliation[2]{organization={School of Statistics, Renmin University of China},
city={Beijing},
postcode={100872},
country={China}}


\begin{abstract}

Automated international classification of diseases (ICD) coding aims to assign multiple disease codes to clinical documents and plays a critical role in healthcare informatics. However, its performance is hindered by the extreme long-tail distribution of the ICD ontology, where a few common codes dominate while thousands of rare codes have very few examples. To address this issue, we propose a Probability-Biased Directed Graph Attention model (ProBias) that partitions codes into common and rare sets and allows information to flow only from common to rare codes. Edge weights are determined by conditional co-occurrence probabilities, which guide the attention mechanism to enrich rare-code representations with clinically related signals. To provide higher-quality semantic representations as model inputs, we further employ large language models to generate enriched textual descriptions for ICD codes, offering external clinical context that complements statistical co-occurrence signals. Applied to automated ICD coding, our approach significantly improves the representation and prediction of rare codes, achieving state-of-the-art performance on three benchmark datasets. In particular, we observe substantial gains in macro-averaged F1 score, a key metric for long-tail classification.

\end{abstract}


\end{frontmatter}

\linenumbers


\nolinenumbers

\section{Introduction}

\begin{figure}[h]
    \centering
    \includegraphics[width=0.8\linewidth]{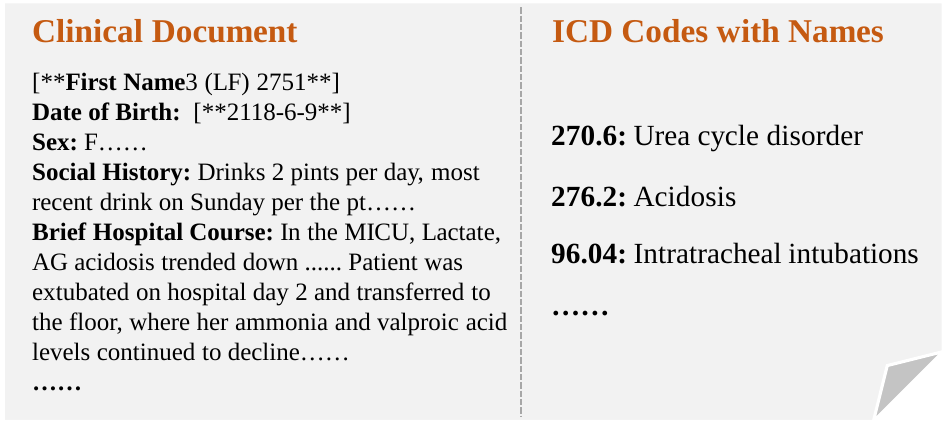}
    \caption{An example of a clinical document assigned with multiple codes ``Urea cycle disorder", ``Acidosis", and others. ``Acidosis" is often triggered by ``Urea cycle disorder", as toxic ammonia buildup disrupts cellular metabolism.}
    \label{fig:example}
\end{figure}

The international classification of diseases (ICD) is a standardized coding system maintained by the World Health Organization (WHO) that assigns unique codes to diagnoses and procedures. Although widely adopted, the manual assignment of ICD codes to clinical documents is labor-intensive, error-prone, and costly, making automated ICD coding an essential tool for modern healthcare systems. Automated ICD coding is typically formulated as a multi-label text classification task \citep{larkey1996combining}, because a single clinical document usually corresponds to multiple diagnoses and procedures (see Figure \ref{fig:example} for example), and therefore must be assigned multiple ICD codes simultaneously.

Automated ICD coding is particularly challenging due to the extreme imbalance in the ICD label space. A small subset of codes, such as those for common chronic conditions, occurs very frequently, whereas the vast majority correspond to rare diseases and appear infrequently. If we view code frequencies as a distribution, this pattern exhibits a long-tail distribution, where a few common codes dominate the dataset while thousands of rare codes appear only a handful of times. For example, in the MIMIC-IV-ICD-10 dataset, over 70\% of the 26,096 distinct codes occur fewer than 10 times \citep{nguyen2023mimic}. This severe long-tail distribution makes it difficult for data-driven models to learn meaningful semantic representations for rare codes, ultimately limiting overall coding performance.

To tackle the multi-label ICD coding problem, some previous studies adopt multi-stage frameworks to handle the ICD system's vast label space. Approaches like GPsoap \citep{yang2023multi} and TwoStage \citep{nguyen2023two} decompose the task into discrete steps, such as candidate code retrieval followed by re-ranking. Another line of research focuses on incorporating external knowledge to enrich code semantics. Methods like MSMN \citep{yuan2022code}, KEPT \citep{yang2022knowledge}, and MSAM \citep{gomes2024accurate} enhance code representations by aligning code names with synonyms from medical ontologies (e.g., Wikidata). While such synonym-based expansion captures lexical variations, it often adds little semantic diversity. For example, the synonyms ``disorder of urea cycle metabolism, unspecified'' and ``disorder of the urea cycle metabolism, nos'' both belong to ICD code 270.6 (``urea cycle disorder''). Consequently, such methods are only marginally effective at injecting meaningful knowledge to enhance code representations, leaving significant room for improvements in classification performance.

Another commonly used strategy is to leverage co-occurrence relationships to capture the complex correlations among codes. Several studies model these correlations using graph networks, motivated by comorbidity as the clinical foundation underlying statistical co-occurrence \citep{valderas2009defining,cao2020hypercore,luo2024corelation,wang2024multi}. However, these approaches often oversimplify co-occurrence as binary connections, ignoring fine-grained information such as the actual co-occurrence probabilities. This limitation reduces the effectiveness of representation learning, particularly for rare codes with scarce training data.

In this work, we propose a method that leverages fine-grained co-occurrence information to improve representation learning for rare ICD codes. We construct a directed bipartite graph encoder with two disjoint node sets (i.e., common codes and rare codes), where edges are directed from common to rare codes and weighted by their conditional co-occurrence probabilities. These probabilities are discretized into bins, which index a learnable probability-biased attention term, a mechanism we refer to as \textit{co-occurrence encoding}. This design allows rare codes to aggregate latent comorbidity information from their frequently observed common counterparts, yielding richer semantic representations. To provide high-quality textual input for the directed bipartite graph encoder, we prompt a medical-domain large language model (LLM) to generate detailed ICD code descriptions that go beyond simple synonyms used in prior work. It can incorporate richer clinical context and comorbidity cues that support more reliable statistical analysis. The resulting code representations are then integrated with a multi-label attention module to produce final predictions. We name our model \textit{ProBias}\footnote{The code for our ProBias model is publicly available at \url{https://github.com/ChTianlEI/ProBias-for-ICD-Coding}.}, derived from its probability-based attention bias. Experiments on three benchmark datasets show that ProBias consistently achieves state-of-the-art performance, delivering substantial improvements on metrics that are particularly sensitive to long-tail label distributions.

The remainder of this paper is organized as follows. Section 2 reviews related work. Section 3 details our methodology. Section 4 presents experiments. Section 5 concludes with limitations and future directions.

\section{Related work}

In this section, we review related work from three perspectives: graph-based models, external knowledge integration, and co-occurrence–driven methods.

\subsection{Graph-based Models}

Graph models play a crucial role in automatic ICD coding. For example, Graph Convolutional Networks (GCNs) \citep{kipf2017semi} have been adopted by ZAGCNN \citep{rios2018few}, MSATT-KG \citep{xie2019ehr}, and HyperCore \citep{cao2020hypercore}. More recently, Transformer-based graph models have gained increasing attention. For example, CoRelation \citep{luo2024corelation} employs a graph Transformer \citep{dwivedi2020generalization} to model the interactions between codes on its constructed simplified graph. Similarly, \citet{wang2024multi} leverage Graphormer \citep{ying2021transformers} to extract the co-occurrence relationships simply based on the connectivity between any two codes. A key innovation of the Transformer-based graph model Graphormer \citep{ying2021transformers} is how it injects structural information into the attention module. Its ``Spatial Encoding" component adds a learnable bias to the attention score between any two nodes. This bias is specifically indexed by the shortest path distance (SPD) between the nodes, allowing the model to incorporate graph topology directly into its calculations. 

\subsection{External Knowledge Integration}

To enhance the semantic representation of ICD codes, a primary strategy involves extending code names with information from external medical knowledge bases. One prominent approach is to extract synonyms. For example, MSMN \citep{yuan2022code} aligns ICD codes with concepts in UMLS to gather synonyms and proposes a multi-synonym attention mechanism where each synonym acts as a separate query to extract relevant text snippets from the clinical note. In addition, MSAM \citep{gomes2024accurate} expands the knowledge sources to include Wikidata and Wikipedia. Recognizing that synonym lists can be repetitive, it also employs a selection algorithm to choose a more representative subset of synonyms for each code. Other works integrate multiple types of knowledge simultaneously. For example, KEPT \citep{yang2022knowledge} integrates a pre-trained language model with three domain-specific knowledge sources: code hierarchy, synonyms, and abbreviations.

\subsection{Co-occurrence Relationships Extraction}

Many existing works utilize co-occurrence among codes to extract their correlations. HyperCore \citep{cao2020hypercore} was the first work to exploit code co-occurrence relationships for automatic ICD coding. It defines the co-occurrence using an adjacency matrix, where the edge weights are directly determined by the raw co-occurrence counts between codes. This structural information is then encoded by a Graph Convolutional Network (GCN). Other approaches have built upon this concept with different strategies for modeling co-occurrence. For instance, \citet{wang2024multi} proposed a method that binarizes co-occurrence relationships by representing the presence or absence of code pairs. These binarized features are then incorporated into a graph transformer model. In another direction, CoRelation \citep{luo2024corelation} integrates the ICD code ontology with the co-occurrence graph to construct a simplified and contextualized graph for each note. To manage complexity, this graph's nodes are reduced by only selecting the top-K codes with the highest initial prediction probabilities for one node set, while using coarse-grained major code categories for the other.

\section{Method}

For automated ICD coding, existing methods commonly treat it as a multi-label text classification problem. Given a clinical document (e.g., discharge summary) $X=\{x_{i}: 1 \leq i \leq T\}$, where $T$ represents the potentially thousands of tokens that compose $X$. Each document $X$ is associated with a ground-truth label set $L = \{l_{i}: 1 \leq i \leq M\}$, where $M$ indicating the number of labels. The objective of automated ICD coding is to predict the label set $L$ for document $X$. Note that, each ICD code represents either a diagnose or a procedure, serving as the task's label, and typically comes with an original name.

\begin{figure}[h]
    \centering
    \includegraphics[width=\linewidth]{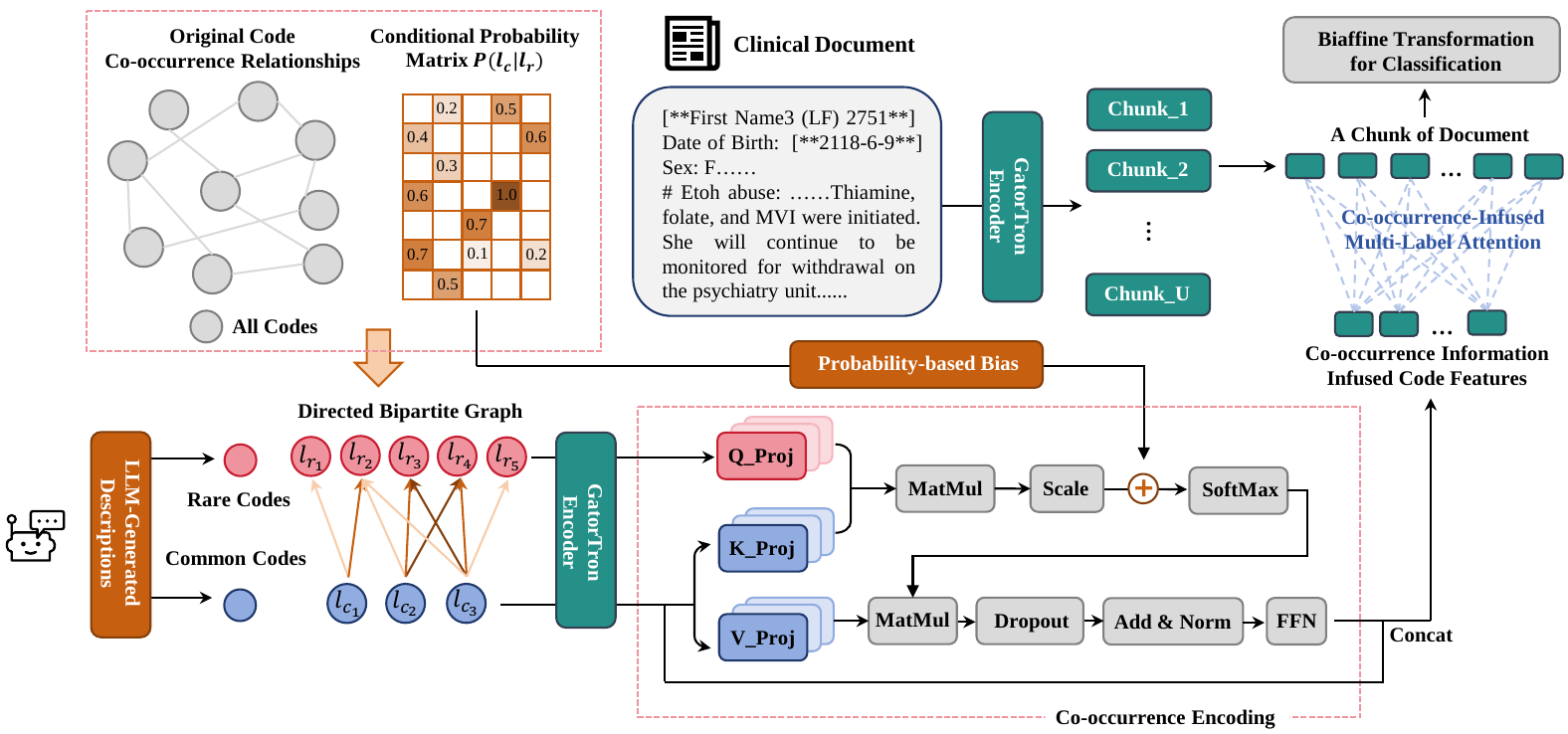}
    \caption{The pipeline of our proposed model, ProBias. The process begins with generating comprehensive code descriptions via a Large Language Model (LLM). Next, a Directed Bipartite Graph is constructed based on the Original Code Co-occurrence Relationships and Conditional Probability Matrix, in which arrows of different colors denote different probabilities. The information flow within this graph is governed by our Co-occurrence Encoding, which injects a learnable bias derived from the Conditional Probability Matrix into the attention scores. The final classification is then performed using a Co-occurrence-Infused Multi-Label Attention mechanism.}
    \label{fig:Model_Structure}
\end{figure}

In this section, we present the technical details of the proposed ProBias model. The overall pipeline is shown in Figure \ref{fig:Model_Structure}. As shown, the method consists of four main components.
First, we generate enriched textual descriptions for each ICD code using LLM, producing high-quality semantic inputs.
Second, we construct a directed bipartite graph based on the original co-occurrence statistics and the conditional probability matrix, where edges are directed from common to rare codes and weighted according to different probability levels.
Third, we introduce a co-occurrence encoding mechanism that enables information to flow through this graph by injecting a learnable probability-based bias into the attention scores, thereby enhancing the representations of data-scarce rare codes.
Finally, we obtain document-level predictions using a co-occurrence-infused multi-label attention module, which aggregates code-aware signals to produce the final ICD assignments.

\subsection{Encoding of Documents and Codes}
To start with, the clinical documents and ICD codes should be converted into embeddings, with documents encoded via a pre-trained language model and codes represented using LLM-enhanced textual descriptions.

\subsubsection{Documents encoding}

We choose GatorTron \citep{yang2022gatortron} as the foundational language model for clinical documents encoding. GatorTron is an encoder model pre-trained on medical corpora, which is publicly available in the NVIDIA NGC\footnote{\url{https://catalog.ngc.nvidia.com/}} Catalog and also through the HuggingFace\footnote{\url{https://huggingface.co/UFNLP/gatortron-base}} library.

As noted in \citep{liu2024lost}, Chunk encoding effectively mitigates the challenges language models encounter when processing lengthy inputs. Given the substantial length of clinical documents, we adopt the approach in \citep{gomes2024accurate}, partitioning each document into $U$ overlapped chunks and processing each chunk independently using GatorTron. This approach circumvents the token limit of the standard Transformer encoder, enabling effective analysis of lengthy clinical documents. Specifically, for a document $X=\{X_1,X_2,...,X_U\}$ that has been divided into $U$ chunks, where each chunk $X_u=\{w_1^u,w_2^u,...,w_T^u\}$ has a fixed length of $T=512$ tokens. We employ the GatorTron model to derive its representation: 
\begin{equation}
  H^{u} = \text{GatorTron}(X_u) \in \mathbb{R}^{T \times d}
\end{equation}
Here, $d$ represents the dimension of the hidden layer.

\subsubsection{Code encoding}
\begin{figure}[h]
    \centering
    \includegraphics[width=\linewidth]{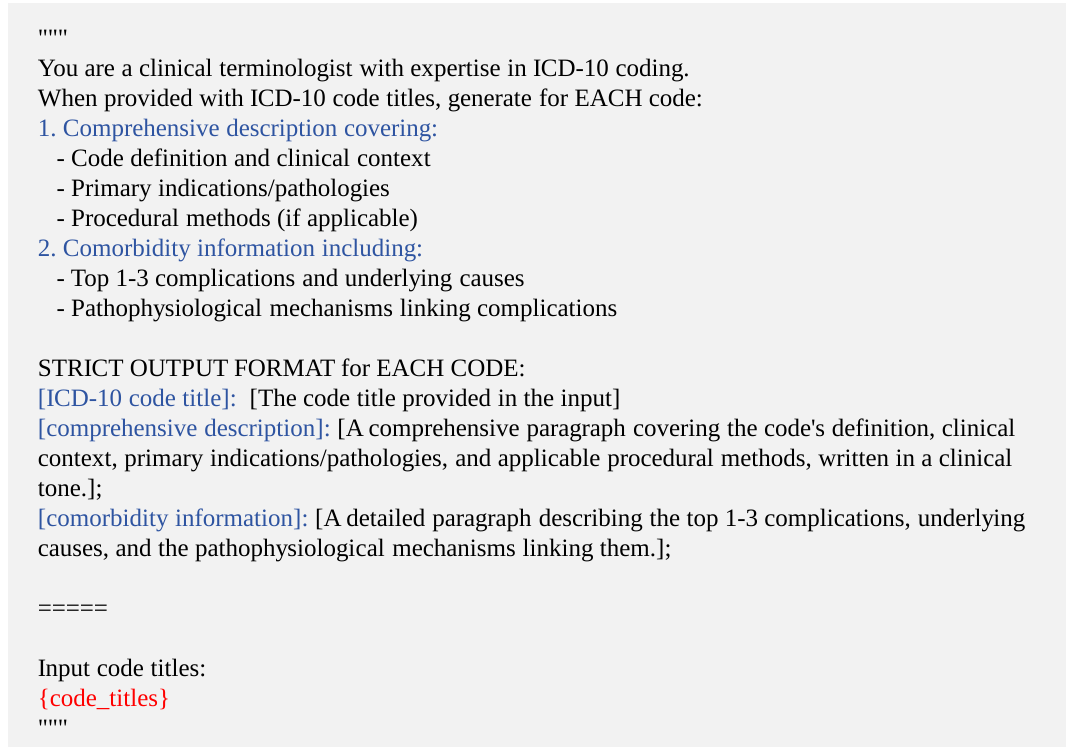}
    \caption{A structured prompt template designed to produce a comprehensive description for each ICD code, covering clinical contexts, procedural methods, and comorbidity information.}
    \label{fig:prompt}
\end{figure}

\begin{figure}[h]
    \centering
    \includegraphics[width=\linewidth]{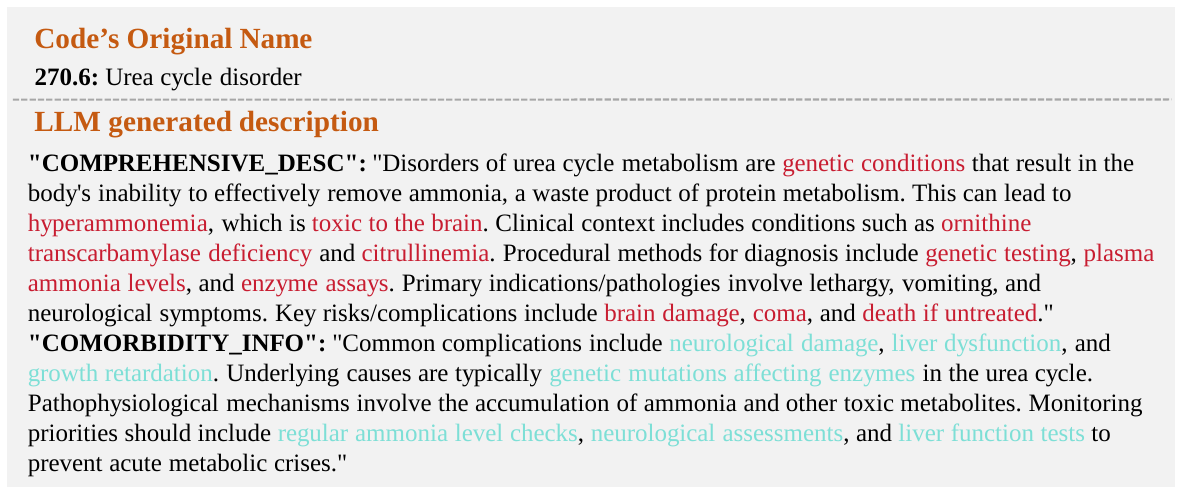}
    \caption{The description of code 270.6 generated by GPT-4o.}
    \label{fig:description}
\end{figure}

For code encoding, we first generate ICD code descriptions using GPT-4o, a state-of-the-art large language model from OpenAI. Its strong semantic understanding, instruction-following ability, and extensive medical knowledge enable the creation of highly structured and clinically accurate descriptions. To ensure consistent and reproducible outputs, we set the temperature parameter to 0.2. The model is guided by a structured prompt template designed to produce comprehensive descriptions for each ICD code, encompassing clinical context, procedural details, and comorbidity information. See Figure \ref{fig:prompt} for the prompt and Figure \ref{fig:description} for an example.

The resulting descriptions are then encoded using the GatorTron model \citep{yang2022gatortron} to obtain the initial code embeddings:
\begin{equation}
      V_i = \text{GatorTron}(d_i)[\text{CLS}]
\end{equation}
Here, $d_i$ denotes the description of the $i$-th code $l_i$, where $i \in \{1,2,...,N\}$ and $N$ is the total number of codes. These initial code embeddings are then input into the directed bipartite graph encoder (as descriped in the next subsection), serving as the basis for capturing fine-grained co-occurrence relationships. Notably, the GatorTron model for these code descriptions remains untrained to mitigate significant GPU memory usage, given the large number of codes.

\subsection{Directed Bipartite Graph Encoder}

In this module, we introduce the Directed Bipartite Graph Encoder for enhancing the representations of rare codes by aggregating latent comorbidity information from their co-occurring common counterparts.

\subsubsection{Graph construction}

As shown by Figure \ref{fig:Model_Structure}, the original code co-occurrence relationships are straightforward: two codes are connected if they co-occur within a document. We calculate a conditional probability metric between rare and common codes:
\begin{equation}
    P(l_{c_j}|l_{r_i})=\frac{N_{l_{r_i}\cap l_{c_j}}}{N_{l_{r_i}}}
\end{equation}
Here, $N_{l_{r_i}\cap l_{c_j}}$ represents the number of times rare code $l_{r_i}$ and common code $l_{c_j}$ co-occur, and $N_{l_{r_i}}$ is the total occurrences of $l_{r_i}$ in the training dataset. The conditional probability matrix is denoted as $\mathbf{P} \in [0,1]^{N_r \times N_c}$, where $N_r$ is the total number of rare codes and $N_c$ is the total number of common codes.

The directed bipartite graph is defined as $\mathcal{G}=(\mathcal{V},\mathcal{E})$, where $\mathcal{V}=\mathcal{V}_{c}\cup\mathcal{V}_{r}$ represents the set of ICD codes, and $\mathcal{V}_{c}\cap\mathcal{V}_{r}=\emptyset$. Here, $\mathcal{V}_{c}=\{l_{c_j}: j \in \{1,2,\cdots,N_c\}\}$ denotes the set of common codes, with $N_c$ being their total number, and $\mathcal{V}_{r}=\{l_{r_i}: i \in \{1,2,\cdots,N_r\}\}$ denotes the set of rare code, with $N_r$ being their total number. The edge set $\mathcal{E}$ consists of directed edges from common codes to rare codes, facilitating information transmission.

\subsubsection{Co-occurrence Encoding with Probability-based Bias}

To effectively model the co-occurrence relationships between ICD codes, we introduce a novel \textit{co-occurrence encoding}. This item is a domain-specific adaptation of the spatial encoding component in the graph transformer model Graphormer \citep{ying2021transformers}.

Transformer layers for sequential data explicitly capture positional dependencies, often via techniques like relative positional encoding \citep{shaw2018self, raffel2020exploring}. However, the graphs lack an inherent sequential order. To address this problem, Graphormer introduces a learnable spatial encoding bias in the attention module, indexed by the shortest path distance (SPD) between nodes. However in the medical domain, the statistical co-occurrence of ICD codes often reflects underlying comorbidity among diseases and provides a more informative signal \citep{valderas2009defining}. Accordingly, our proposed bias is indexed not by distance but by the conditional probability between code pairs, which enables the model to capture a finer-grained, representation-level understanding of co-occurrence relationships. The implementation of co-occurrence encoding is implemented by discretizing continuous probability values into a fixed number of bins. The bin index $\varphi(l_{r_i},l_{c_j})$ for any code pair is generated via quantile binning, as detailed in Algorithm \ref{alg:quantile_binning}.

\begin{algorithm}[h]
\small
\DontPrintSemicolon
\SetAlgoVlined
\SetKwInOut{Input}{Input}
\SetKwInOut{Output}{Output}
\caption{Binning for Probability-based Bias}
\label{alg:quantile_binning}

\Input{Conditional Probability Matrix $\mathbf{P} \in \mathbb{R}^{N_r \times N_c}$; Bin number $B$.}
\Output{Bin indices $\varphi(l_{r_i},l_{c_j}) \in \{0,1,...,B\}, i\in \{1 ,..., N_r\}, j\in \{1 ,..., N_c\}$}

\textbf{Step 1: Quantile Binning with Zero Handling}
\Begin{
    1. \textbf{Flatten} matrix: $\mathbf{P}_{flat} \gets \mathbf{P}.flatten()$\;
    2. \textbf{Separate zeros}: 
    $\mathbf{P}_{nonzero} \gets \mathbf{P}_{flat}[\mathbf{P}_{flat} \neq 0]$\;
    3. \textbf{Compute quantiles}:
    \Begin{
        $\text{quantiles} \gets [0.0, 0.1, ..., 1.0]$\;
        $\text{bins} \gets \text{np.quantile}(\mathbf{P}_{nonzero}, \text{quantiles})$\;
    }
    4. \textbf{Merge boundaries}:
    $\text{bins} \gets \text{np.concatenate}(([0.0], \text{bins}))$\;
}

\textbf{Step 2: Index Assignment}
\Begin{
    \For{$i \leftarrow 1$ \KwTo $N_r$}{
        \For{$j \leftarrow 1$ \KwTo $N_c$}{
            \uIf{$\mathbf{P}(i,j) = 1$}{
                $\varphi(l_{r_i}, l_{c_j}) \leftarrow B$\;
            }
            \Else{
                 $\varphi(l_{r_i}, l_{c_j}) \leftarrow \text{torch.bucketize}(\mathbf{P}(i,j), \text{bins}, \text{right=True})-1$\;
            }
        }
    }
}
\textbf{return} $\varphi$
\end{algorithm}

The probability-based bias $c_{\varphi(l_{r_i},l_{c_j})}$ is a learnable scalar indexed directly by $\varphi(l_{r_i},l_{c_j})$. This bias is then incorporated into the attention calculation for the $k$-th head ($k \in {1,...,K}$, where $K$ is the number of heads) between a rare code $l_{r_i}$ and a common code $l_{c_j} $as follows:

\begin{equation}
\begin{aligned}
A_{l_{r_{i}}l_{c_{j}}}^{k} = \underset{l_{c_j} \in \mathcal{V}_c}{\text{SoftMax}} \left( \frac{(V_{l_{r_i}} W_R^k)(V_{l_{c_j}} W_C^k)^{\prime}}{\sqrt{d_k}} + c_{\varphi(l_{r_i},l_{c_j})}^{k} + m_{l_{r_i},l_{c_j}} \right)
\end{aligned}
\end{equation}
Here, $W_R$ and $W_C$ are the projection matrices for the query and key of the $k$-th attention head, respectively, and $d_k$ is the dimension of the query and key for the $k$-th head ($d_k=d/K$). The probability-based bias for the $k$-th head is denoted as $c^k_{\varphi(l_{r_i},l_{c_j})}$. The term $m_{l_{r_i},l_{c_j}}$ serves as a binary mask to control co-occurrence. Specifically, it is set to $-\infty$ if codes $l_{r_i}$ and $l_{c_j}$ do not co-occur, effectively preventing attention between them. Conversely, it is set to 0 if they do co-occur, allowing common codes to transmit information to their co-occurring rare codes.

The output of the multi-head attention mechanism for the rare code $l_{r_i}$ is obtained by concatenating the outputs from all attention heads, followed by a linear transformation:
\begin{equation}
Att_{l_{r_{i}}} = W_{att} \cdot \text{Concat}(head_1, head_2, ..., head_H)
\end{equation}
Here $head_h=\sum_j A^h_{l_{r_{i}}l_{c_{j}}}V_{l_{c_j}}W^h_V$, $W_V$ is the projection matrix for $l_{c_{j}}$, and $W_{att}$ is the linear layer that combines the outputs of all heads. The resulting rare code attention features, $Att_{l_{r_{i}}}$, along with common code features, $V_{l_{c_j}}$, are then processed through a dropout layer, normalization, residual connection, and MLP layers to generate the final graph encoder output, $V^G=\{V_i^G\}$, where $i \in \{1,2,...,N\}$ and $N$ is the total number of codes. This feed-forward layer performs local refinement on all codes, which enhances the discriminative power of the output features and improves classification recognition capability.

\subsection{Co-occurrence-Infused Multi-Label Attention}

The multi-label attention mechanism is a prevalent technique in text classification and has been adapted by researchers for various task designs. Notable examples include \textit{hierarchy-aware multi-label attention} \citep{zhou2020hierarchy} and \textit{multi-synonyms attention} \citep{yuan2022code,gomes2024accurate}, which in turn draw inspiration from the multi-head attention mechanism of the transformer architecture \citep{vaswani2017attention}. To integrate the chunked document representation $H=\{H^u\}$ with the comorbidity information infused code features $V^G = \{V_i^G\}$ from the graph encoder, we implement a multi-label attention mechanism to extract text features specific to each code. While our implementation utilizes multi-head attention, for clarity of presentation, the following equations illustrate the computation for a single head. First, let the document chunk representation $H^{u} \in \mathbb{R}^{T \times d}$ be a sequence of token representations $H^u = \{h_1^u, h_2^u,...,h_T^u\}$, where $h_T^u \in \mathbb{R}^d$ is the representation for the $t$-th token. The code features $V^G_i$ are then used to query this token sequence as follows:
\begin{equation}
{\alpha}_{u}^{i} = \underset{t \in {1,...,T}}{\text{SoftMax}} \left( (W_Q V^G_i) \cdot \text{Tanh}(W_K h_t^u)^T \right)
\end{equation}
Here, $W_Q$ and $W_K$ are projection matrices, and ${\alpha}_{u}^{i} \in \mathbb{R}^{T}$ quantifies the attention of code $l_i$ across the tokens of the $u$-th document chunk $X_u$. We leverage this attention distribution as weights to compute a code-specific representation for the current chunk, i.e.,
\begin{equation}
    R_u^i = \sum_{t=1}^{T} \alpha_{u,t}^i H_t^u
\end{equation}

For classification of the document $H$, we employ a biaffine transformation with the derived chunk-level representation $R_{u}^{i}$:
\begin{equation}
    \hat{y}_i = \sigma(\text{MaxPool}_u(R_u^{i}WV_i^G)).
\end{equation}
Here, the $\text{MaxPool}_u$ operation aggregates representations across all $U$ chunks into a unified document-level score and $\sigma$ is the sigmoid activation, providing the probability $\hat{y}_i$ for code $l_i$.



Finally, for model training, we employ the widely-used binary cross-entropy (BCE) loss. This loss function is averaged over all samples within a batch. For a single document, the BCE loss is defined as:
\begin{equation}
\mathcal{L}_{BCE} = -\frac{1}{N} \sum_{i=1}^{N} [y_{i} \log(\hat{y}_{i}) + (1-y_{i})\log(1-\hat{y}_{i}) ]
\end{equation}
where $N$ is the total number of codes.

\section{Experiments}
\subsection{Dataset}
Our experiments are conducted on the full versions of three benchmark datasets: MIMIC-III-ICD-9, MIMIC-IV-ICD-9, and MIMIC-IV-ICD-10. These datasets are derived from the MIMIC-III \citep{johnson2016mimiciii} and MIMIC-IV \citep{johnson2023mimiciv_note} clinical databases, which we access via PhysioNet\footnote{\url{https://physionet.org/}} after completing the required ethical training program. We adopt the widely-used data splits and processing procedures in \citep{mullenbach2018explainable} for MIMIC-III, and \citep{nguyen2023mimic} for the two MIMIC-IV datasets, ensuring a comprehensive evaluation on the complete and long-tail label space. The summary statistics of the three datasets can be found in Tables \ref{tab:mimic-iii-icd9-stats}, \ref{tab:mimic-iv-icd9-stats}, and \ref{tab:mimic-iv-icd10-stats}.

\begin{table}[h]
\centering
\caption{Statistics of the MIMIC-III-ICD-9 dataset.}
\begin{tabular}{lcccc}
\toprule
\multirow{2}{*}{\textbf{Statistics}} & \multicolumn{4}{c}{\textbf{MIMIC-III-ICD-9}} \\
\cline{2-5}
 & \textbf{All} & \textbf{Train} & \textbf{Dev} & \textbf{Test} \\
\hline
\textbf{Number of Documents.} & 52,726 & 47,723 & 1,631 & 3,372 \\
\textbf{Average Number of Tokens per Document.} & 2,768 & 2,707 & 3,338 & 3,360 \\
\textbf{Average Number of Codes per Document.} & 15.9 & 15.7 & 17.4 & 18.0 \\
\textbf{Total Number of Codes.} & 8,921 & 8,685 & 3,009 & 4,075 \\
\bottomrule
\end{tabular}

\label{tab:mimic-iii-icd9-stats}
\end{table}

\begin{table}[h]
\centering
\caption{Statistics of the MIMIC-IV-ICD-9 dataset.}
\begin{tabular}{lcccc}
\toprule
\multirow{2}{*}{\textbf{Statistics}} & \multicolumn{4}{c}{\textbf{MIMIC-IV-ICD-9}} \\
\cline{2-5}
 & \textbf{All} & \textbf{Train} & \textbf{Dev} & \textbf{Test} \\
\hline
\textbf{Number of Documents.} & 209,352 & 188,533 & 7,110 & 13,709 \\
\textbf{Average Number of Tokens per Document.} & 1,693 & 1,693 & 1,708 & 1,695 \\
\textbf{Average Number of Codes per Document.} & 13.3 & 13.3 & 13.5 & 13.3 \\
\textbf{Total Number of Codes.} & 11,331 & 11,145 & 5,115 & 6,264 \\
\bottomrule
\end{tabular}
\label{tab:mimic-iv-icd9-stats}
\end{table}

\begin{table}[!h]
\centering
\caption{Statistics of the MIMIC-IV-ICD-10 dataset.}
\begin{tabular}{lcccc}
\toprule
\multirow{2}{*}{\textbf{Statistics}} & \multicolumn{4}{c}{\textbf{MIMIC-IV-ICD-10}} \\
\cline{2-5}
 & \textbf{All} & \textbf{Train} & \textbf{Dev} & \textbf{Test} \\
\hline
\textbf{Number of Documents.} & 122,309 & 110,441 & 4,017 & 7,851 \\
\textbf{Average Number of Tokens per Document.} & 1,956 & 1,958 & 1,968 & 1,931 \\
\textbf{Average Number of Codes per Document.} & 16.1 & 16.1 & 16.2 & 15.8 \\
\textbf{Total Number of Codes.} & 26,096 & 25,230 & 6,738 & 9,159 \\
\bottomrule
\end{tabular}
\label{tab:mimic-iv-icd10-stats}
\end{table}

\begin{figure}[h]
    \centering
    \begin{subfigure}[b]{0.33\linewidth}
        \includegraphics[width=0.95\linewidth]{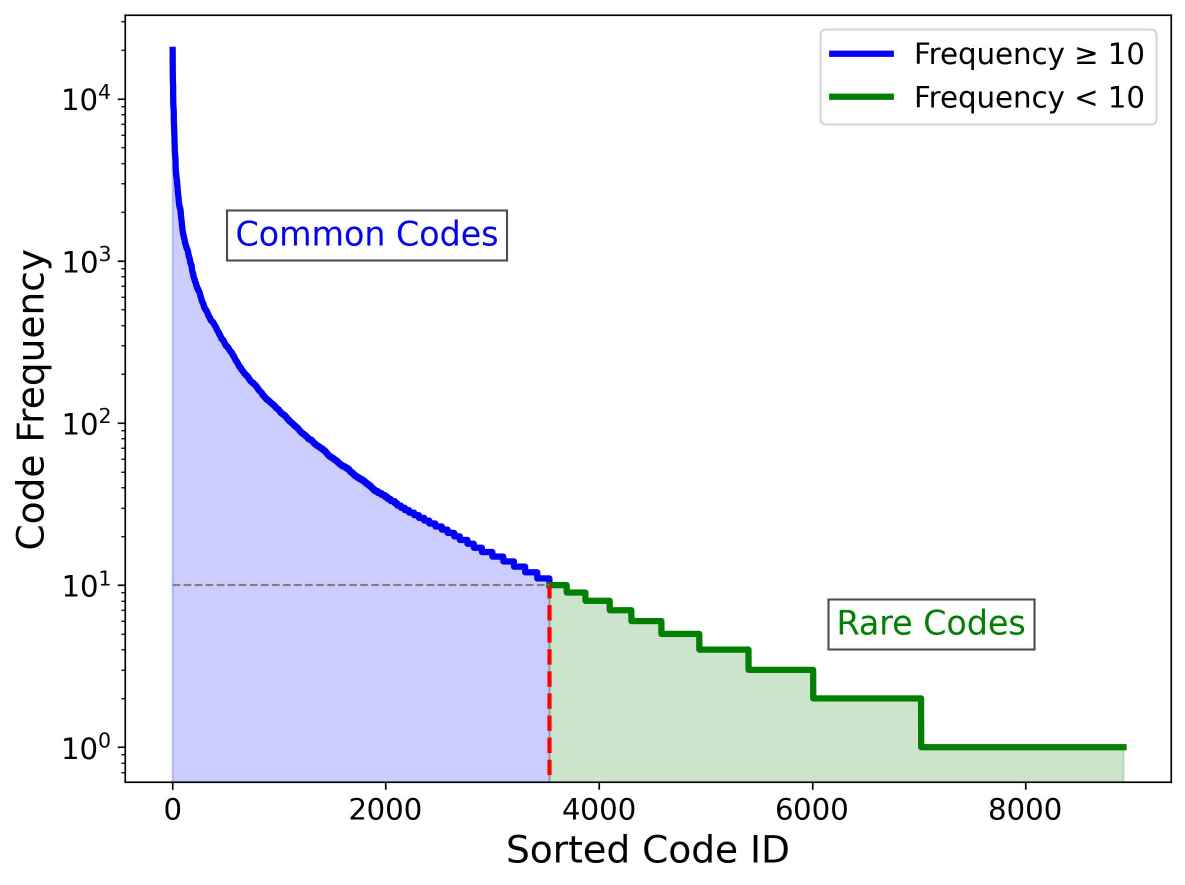}
        \caption{MIMIC-III-ICD-9}
    \end{subfigure}
    \hspace{-3mm}
    \begin{subfigure}[b]{0.33\linewidth}
        \includegraphics[width=0.95\linewidth]{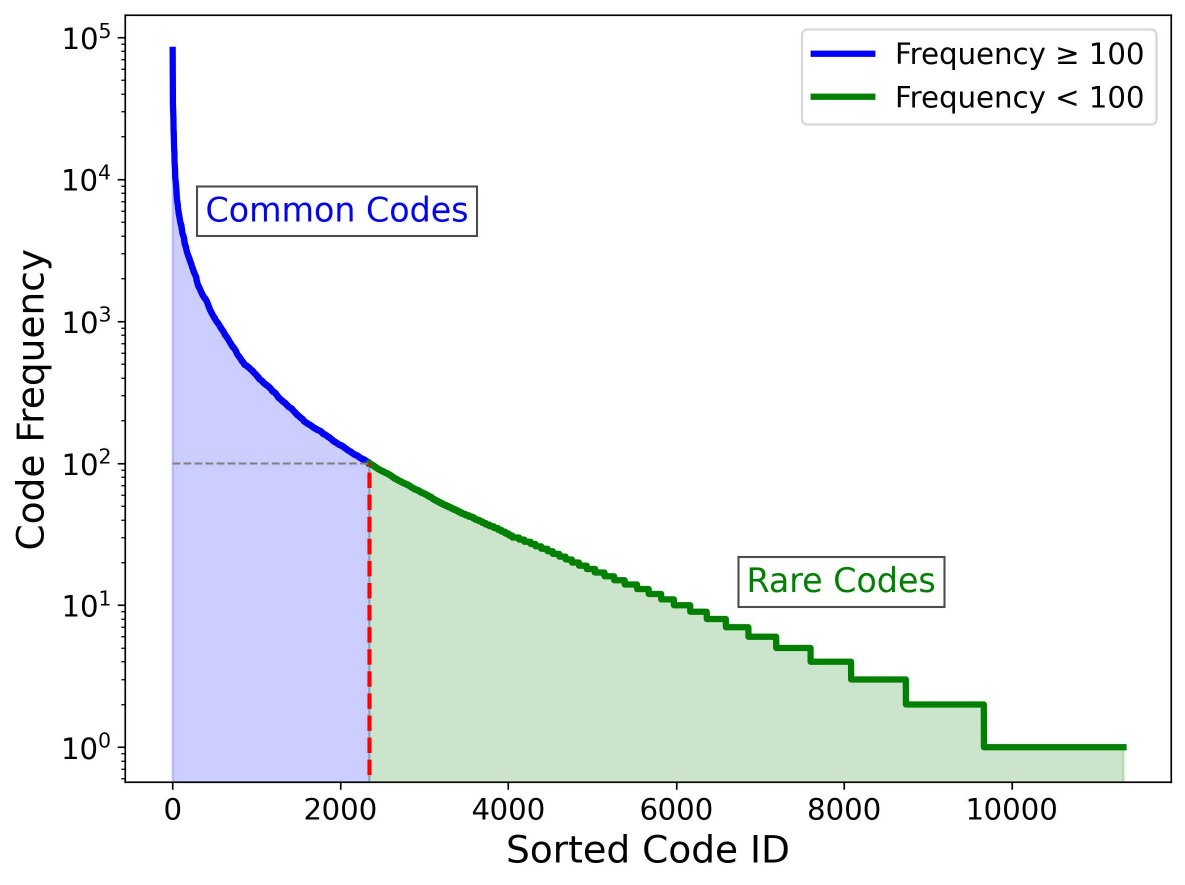}
        \caption{MIMIC-IV-ICD-9}
    \end{subfigure}
    \hspace{-3mm}
    \begin{subfigure}[b]{0.33\linewidth}
        \includegraphics[width=0.95\linewidth]{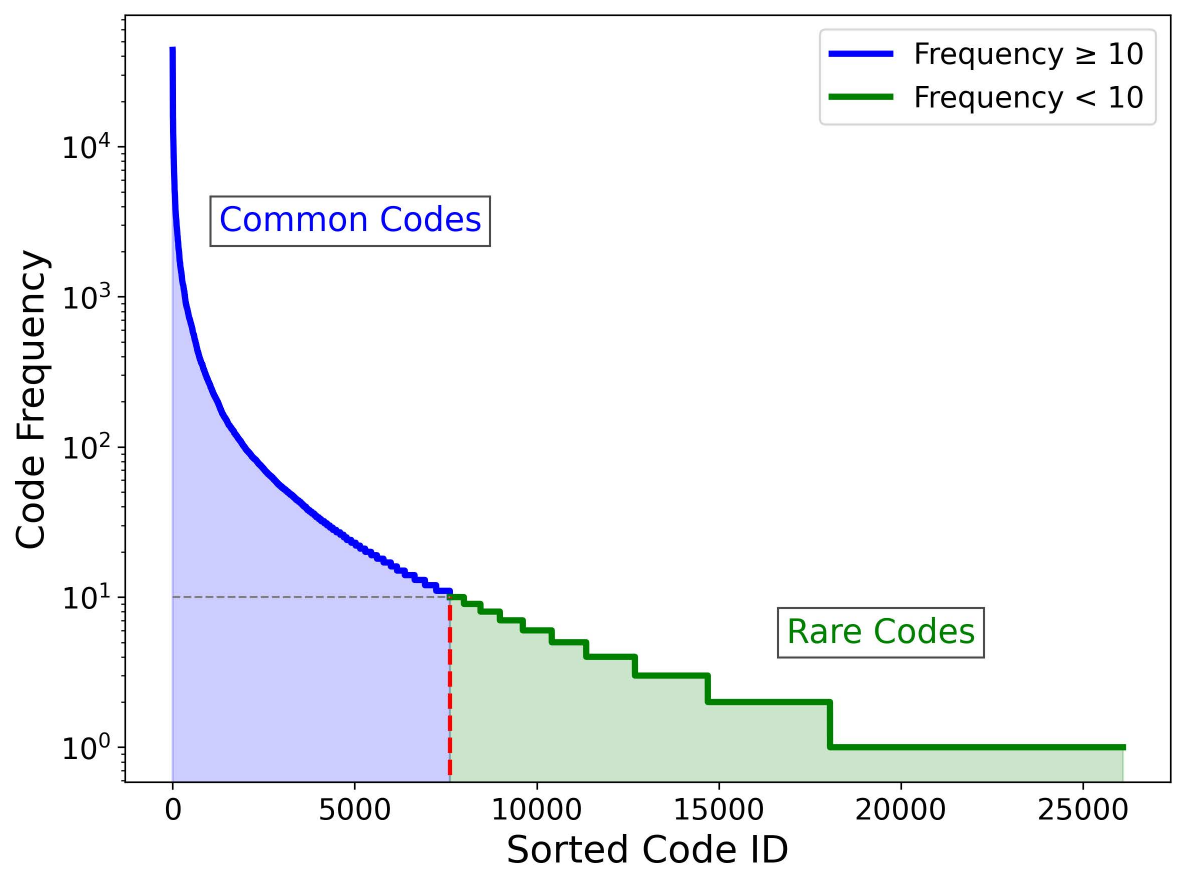}
        \caption{MIMIC-IV-ICD-10}
    \end{subfigure}
    \caption{The long-tail label distributions of three benchmark datasets. ICD codes are sorted in descending order according to their frequency, measured by the number of associated documents.}
    \label{fig:MIMICdistribution}
\end{figure}

In the three datasets, we categorize codes as ``rare" based on their frequency in the training data. The thresholds are set to fewer than 10 occurrences for MIMIC-III-ICD-9 and MIMIC-IV-ICD-10, and fewer than 100 for MIMIC-IV-ICD-9. This results in a substantial portion of codes (approximately 70\%) being categorized as rare (see Figure \ref{fig:MIMICdistribution}). Our segmentation strategy follows prior studies in other domains that address long-tail label distributions \citep{wang2017learning,liu2019large,xiao2021does,wang2024value}.

\subsection{Baselines and Evaluation Metrics}
We compare our model with existing outstanding methods.
\begin{itemize}
    \item CAML: It aggregates information across the document through a convolutional neural network and an attention module \citep{mullenbach2018explainable}.
\item LAAT: It handles both the various lengths and the interdependence of the ICD code related text fragments \citep{vu2021label}.
\item JointLAAT: It minimizes the joint losses of both parent and child codes from code ontology to address long-tail distribution \citep{vu2021label}.
\item MSMN: It proposes a multiple synonyms matching network to leverage synonyms for better code representation learning \citep{yuan2022code}.
\item PLM-ICD: It develops a framework for automatic ICD coding with pretrained language models \citep{huang2022plm}.
\item CoRalation: It employs a dependent learning paradigm that considers the context of clinical notes in modeling all possible code relations \citep{luo2024corelation}.
\item MSAM: It employs a selection algorithm to choose a more distinct subset of synonyms for each code \citep{gomes2024accurate}.
\end{itemize}

Following these studies, we evaluate our model ProBias using a comprehensive set of metrics: Macro AUC, Micro AUC, Macro F1, Micro F1, and Precision@8.
\begin{itemize}
\item Macro AUC and Macro F1 are computed by first calculating the metric for each class independently and then averaging across all classes. This treats all codes equally, regardless of their frequency, and is therefore particularly sensitive to performance on rare codes in the long-tail distribution.

\item Micro AUC and Micro F1, in contrast, aggregate predictions across all instances before computing the metric, which gives more weight to frequent codes and reflects overall performance on the dataset.

\item Precision@8 measures the fraction of correctly predicted ICD codes among the top 8 predictions for each document, evaluating the model’s ability to prioritize the most relevant codes.
\end{itemize}

We place special emphasis on Macro F1, as it directly captures performance on rare classes, which are often the most challenging in long-tail ICD coding.

\subsection{Implementation Details}

We conduct all experiments on a single NVIDIA H100 GPU with 80GB of memory. Our model's training hyper-parameters remain consistent across all three datasets. For input processing, we set the maximum input length to 6122 tokens, dividing them into 512-token chunks with an overlapping window of 255. The GatorTron's hidden size is 1,024. The attention mechanism within the graph utilizes a hidden size of 512 and 4 heads. The feed-forward layer includes a dropout rate of 0.1, and the linear layer in it also has a hidden size of 512. To enhance computational efficiency and reduce GPU memory consumption, we train our model using BFloat16 precision. We optimize our model with AdamW, employing a starting learning rate of $2\times 10^{-5}$ and a linear decay schedule. To prevent overfitting, early stopping is applied based on the Macro F1 score on the validation set. Our model is trained for 15 epochs with a batch size of 1 and 16 gradient accumulation steps.

\subsection{Main Results}
As demonstrated in Table \ref{tab:mimic-iii-icd9}, \ref{tab:mimic-iv-icd9}, and \ref{tab:mimic-iv-icd10}, our model, ProBias, achieves new state-of-the-art (SOTA) results across all three benchmark datasets. On MIMIC-III-ICD-9, ProBias substantially outperforms other methods, boosting the Macro F1 score from 11.4 to 14.2 (a 2.8-point improvement). This superiority continues on the more recent MIMIC-IV datasets. Specifically, on MIMIC-IV-ICD-9, ProBias improves the SOTA Macro F1 score from 15.9 to 18.6, and on MIMIC-IV-ICD-10, it lifts the Macro F1 from 6.3 to 7.7.

\begin{table}[h]
\centering
\caption{Performance comparison with baseline models on MIMIC-III-ICD-9 dataset. Our ProBias results represent the average of five runs with different random seeds. Baseline results are taken from their original publications, with the exception of MSAM, which we reproduced using the public code. The best results are highlighted in \textbf{Bold}.}
\begin{tabular}{lccccc}
\toprule
\multirow{2}{*}{\textbf{Method}} & \multicolumn{2}{c}{\textbf{AUC}} & \multicolumn{2}{c}{\textbf{F1}} & \textbf{Pre} \\
\cline{2-6}
 & \textbf{Macro} & \textbf{Micro} & \textbf{Macro} & \textbf{Micro} & \textbf{P@8} \\
 \hline
\textbf{CAML} \citep{mullenbach2018explainable}     & 89.5        & 98.6        & 8.8        & 53.9       & 70.9 \\
\textbf{LAAT} \citep{vu2021label}      & 91.9        & 98.8        & 9.9        & 57.5       & 73.8 \\
\textbf{JointLAAT} \citep{vu2021label} & 92.1        & 98.8        & 10.7       & 57.5       & 73.5 \\
\textbf{MSMN} \citep{yuan2022code}      & 95.0        & \textbf{99.2}    & 10.3       & 58.4       & 75.2 \\
\textbf{PLM-ICD} \citep{huang2022plm}   & 92.6        & 98.9        & 10.4       & 59.8       & \textbf{77.1} \\
\textbf{CoRalation} \citep{luo2024corelation} & 95.2        & \textbf{99.2}    & 10.2       & 59.1       & 76.2 \\
\textbf{MSAM}   \citep{gomes2024accurate}    & 94.6        & 99.1        & 11.4       & 59.2       & 75.9 \\
\hline
\textbf{ProBias (ours)}   & \textbf{95.3}    & \textbf{99.2}    & \textbf{14.2}   & \textbf{60.8}   & 77.0 \\
\bottomrule
\end{tabular}

\label{tab:mimic-iii-icd9}
\end{table}

\begin{table}[h]
\centering
\caption{Performance comparison with baseline models on MIMIC-IV-ICD-9 dataset. Our ProBias results represent the average of five runs with different random seeds. Baseline results are from \citet{luo2024corelation}, with the exception of MSAM, which we reproduced using the public code. The best results are highlighted in \textbf{Bold}.}
\begin{tabular}{lccccc}
\toprule
\multirow{2}{*}{\textbf{Method}} & \multicolumn{2}{c}{\textbf{AUC}} & \multicolumn{2}{c}{\textbf{F1}} & \textbf{Pre} \\
\cline{2-6}
 & \textbf{Macro} & \textbf{Micro} & \textbf{Macro} & \textbf{Micro} & \textbf{P@8} \\
 \hline
\textbf{CAML} \citep{mullenbach2018explainable}     & 93.5        & 99.3        & 11.1       & 57.3       & 64.9 \\
\textbf{LAAT} \citep{vu2021label}      & 95.2        & 99.5        & 13.1       & 60.3       & 67.5 \\
\textbf{JointLAAT} \citep{vu2021label} & 95.6        & 99.5        & 14.2       & 60.4       & 67.5 \\
\textbf{MSMN} \citep{yuan2022code}      & 96.8        & \textbf{99.6}    & 13.9       & 61.2       & 68.9 \\
\textbf{PLM-ICD} \citep{huang2022plm}   & 96.6        & 99.5        & 14.4       & 62.5       & 70.3 \\
\textbf{CoRalation} \citep{luo2024corelation} & 96.8        & 99.5        & 15.0       & 62.4       & 70.1 \\
\textbf{MSAM}   \citep{gomes2024accurate}    & \textbf{97.5}    & \textbf{99.6}    & 15.9       & \textbf{63.2}   & \textbf{71.0} \\
\hline
\textbf{ProBias (ours)}  & \textbf{97.5}    & \textbf{99.6}    & \textbf{18.6}   & \textbf{63.2}   & 70.5 \\
\bottomrule
\end{tabular}

\label{tab:mimic-iv-icd9}
\end{table}

\begin{table}[h]
\centering
\caption{Performance comparison with baseline models on MIMIC-IV-ICD-10 dataset. Our ProBias results represent the average of five runs with different random seeds. Baseline results are from \citet{luo2024corelation}, with the exception of MSAM, which we reproduced using the public code. The best results are highlighted in \textbf{Bold}.}
\begin{tabular}{lccccc}
\toprule
\multirow{2}{*}{\textbf{Method}} & \multicolumn{2}{c}{\textbf{AUC}} & \multicolumn{2}{c}{\textbf{F1}} & \textbf{Pre} \\
\cline{2-6}
 & \textbf{Macro} & \textbf{Micro} & \textbf{Macro} & \textbf{Micro} & \textbf{P@8} \\
 \hline
\textbf{CAML} \citep{mullenbach2018explainable}     & 89.9        & 98.8        & 4.1        & 52.7       & 64.4 \\
\textbf{LAAT} \citep{vu2021label}      & 93.0        & 99.1        & 4.5        & 55.4       & 67.0 \\
\textbf{JointLAAT} \citep{vu2021label} & 93.6        & 99.3        & 5.7        & 55.9       & 66.9 \\
\textbf{MSMN} \citep{yuan2022code}      & 97.1        & 99.6        & 5.4        & 55.9       & 67.7 \\
\textbf{PLM-ICD} \citep{huang2022plm}   & 91.9        & 99.0        & 4.9        & 57.0       & 69.5 \\
\textbf{CoRalation} \citep{luo2024corelation} & 97.2        & 99.6        & 6.3        & 57.8       & 70.0 \\
\textbf{MSAM}   \citep{gomes2024accurate}    & \textbf{97.7}    & \textbf{99.7}    & 6.3        & 58.5       & \textbf{70.4} \\
\hline
\textbf{ProBias (ours)}   & 97.3        & 99.6        & \textbf{7.7}    & \textbf{58.7}   & 70.0 \\
\bottomrule
\end{tabular}

\label{tab:mimic-iv-icd10}
\end{table}

The substantial improvements in Macro F1 across all datasets are particularly noteworthy. This metric is crucial for evaluating performance on long-tail distributions, as it is highly sensitive to a model's effectiveness on rare classes, thus validating our model's superior ability to handle the long-tail challenge. Concurrently, strong results in both Micro F1 and Precision@8 confirm the model's high overall accuracy and its capacity to rank the most relevant labels effectively.

\subsection{Analysis of Key Hyperparameters}
We use the MIMIC-III-ICD-9 dataset as an example to determine the optimal settings for key hyperparameters, specifically the number of bins for our co-occurrence encoding and the number of layers in the directed bipartite graph encoder. The hyperparameter selection procedure is similar for the other two datasets.

As shown in Table \ref{tab:mimiciii_bin}, we evaluate bin numbers of 5, 10, and 15. We select 10 bins, as it achieves the highest Macro F1 score (14.2), providing a good balance between information granularity and model generalization. For the number of graph layers (see Table \ref{tab:mimiciii_layer}), we adopt a single-layer architecture. Although a 2-layer model yields a slightly higher Macro F1 (14.4 vs. 14.2), the single-layer configuration is substantially more efficient, which is important for scalability to larger datasets such as MIMIC-IV-ICD-10. In summary, for all experiments, we use 10 bins for co-occurrence encoding and a single-layer directed bipartite graph encoder, which together provide a strong balance of performance and computational efficiency.

\begin{table}[h]
\centering
\caption{Performance comparison of different bin number in \textit{co-occurrence encoding} (5, 10 and 15) on MIMIC-III-ICD-9 dataset. The best results are highlighted in \textbf{Bold}.}
\label{tab:mimiciii_bin}
\begin{tabular}{lccccccc}
\toprule
\multirow{2}{*}{\textbf{Models}}
 & \multicolumn{2}{c}{\textbf{AUC}} & \multicolumn{2}{c}{\textbf{F1}} & \multicolumn{3}{c}{\textbf{Pre}}\\
\cline{2-8}
 & \textbf{Macro} & \textbf{Micro} & \textbf{Macro} & \textbf{Micro} & \textbf{P@5} & \textbf{P@8} & \textbf{P@15}\\
\hline
\textbf{ProBias ($b_5$)} & \textbf{95.6} & \textbf{99.3} & 13.7 & 60.5 & 83.5 & 76.6 & 61.9 \\

\textbf{ProBias ($b_{10}$)} & 95.3 & 99.2 & \textbf{14.2} & \textbf{60.8} & \textbf{84.0} & \textbf{77.0} & \textbf{62.0} \\

\textbf{ProBias ($b_{15}$)} & 95.3 & 99.2 & 13.8 & 60.3 & 83.4 & 76.6 & 61.6 \\

\bottomrule

\end{tabular}

\end{table}

\begin{table}[h]
\centering
\caption{Performance comparison of different layer number in \textit{directed bipartite graph encoder} (1, 2 and 3) on MIMIC-III-ICD-9 dataset. The best results are highlighted in \textbf{Bold}.}
\begin{tabular}{lccccccc}
\toprule
\multirow{2}{*}{\textbf{Models}}
 & \multicolumn{2}{c}{\textbf{AUC}} & \multicolumn{2}{c}{\textbf{F1}} & \multicolumn{3}{c}{\textbf{Pre}}\\
\cline{2-8}
 & \textbf{Macro} & \textbf{Micro} & \textbf{Macro} & \textbf{Micro} & \textbf{P@5} & \textbf{P@8} & \textbf{P@15}\\
\hline
\textbf{ProBias ($l_1$)} & 95.3 & 99.2 & 14.2 & \textbf{60.8} & \textbf{84.0} & \textbf{77.0} & \textbf{62.0} \\

\textbf{ProBias ($l_2$)} & 95.3 & 99.2 & \textbf{14.4} & 60.5 & 83.4 & 76.7 & 61.8 \\

\textbf{ProBias ($l_3$)} & \textbf{95.5} & \textbf{99.3} & 14.3 & 60.6 & 83.6 & 76.7 & 61.9 \\

\bottomrule

\end{tabular}

\label{tab:mimiciii_layer}
\end{table}

\subsection{Ablation Study}
We conduct ablation experiments on the MIMIC-III-ICD-9 dataset to evaluate the contribution of each module. The model variants considered in the ablation study are as follows:
\begin{itemize}
    \item ProBias (MI) serves as the baseline, using a simple adjacency graph with Mutual Information flow between all co-occurring codes.
\item ProBias (DI) improves upon this by introducing Directional Information flow from common to co-occurring rare codes.

\item ProBias (DI+CE) incorporates our key contribution, replacing the directional flow with fine-grained Co-occurrence Encoding that injects a probability-based bias.

\item ProBias (DI+CE+LLM), the full model, further leverages LLM-generated code descriptions instead of plain code names as graph input, achieving the best performance.
\end{itemize}
These ablations demonstrate the incremental benefit of each component in enhancing rare-code representation and overall ICD coding accuracy.

The detailed results are summarized in Table \ref{tab:ablation}). As shown, the base version, ProBias(MI), already provides a strong performance foundation, reflecting the strength of the overall pipeline. Building upon this, the incremental improvements observed from ProBias(MI) to ProBias(DI) demonstrate the benefit of introducing directional information flow, allowing rare codes to selectively aggregate information from co-occurring common codes. Adding Co-occurrence Encoding in ProBias(DI+CE) further refines this process by injecting fine-grained, probability-based biases into the attention mechanism, enhancing the representation of rare codes in a statistically informed manner. Finally, ProBias(DI+CE+LLM) leverages LLM-generated code descriptions, enriching the graph node embeddings with detailed clinical context and comorbidity knowledge, which provides the most significant boost in performance.

\begin{table}[h]
\centering
    \caption{Performance comparison of diffenrent ablation versions. The best results are highlighted in \textbf{Bold}.}
\begin{tabular}{lccccccc}
\toprule
\multirow{2}{*}{\textbf{Models}}
 & \multicolumn{2}{c}{\textbf{AUC}} & \multicolumn{2}{c}{\textbf{F1}} & \multicolumn{3}{c}{\textbf{Pre}}\\
\cline{2-8}
 & \textbf{Macro} & \textbf{Micro} & \textbf{Macro} & \textbf{Micro} & \textbf{P@5} & \textbf{P@8} & \textbf{P@15}\\
\hline
\textbf{ProBias (MI)} & 95.5 & 99.2 & 11.8 & 60.0 & 83.2 & 76.1 & 61.4\\
\textbf{ProBias (DI)} & 95.6 & \textbf{99.3} & 12.4 & 60.4 & 84.0 & 76.9 & 61.7 \\
\textbf{ProBias (DI+CE)} & \textbf{95.7} & \textbf{99.3} & 13.3 & 60.7 & \textbf{84.3} & \textbf{77.2} & \textbf{62.0}\\ 
\textbf{ProBias (DI+CE+LLM)} & 95.3 & 99.2 & \textbf{14.2} & \textbf{60.8} & 84.0 & 77.0 & \textbf{62.0} \\
\bottomrule
\end{tabular}

    \label{tab:ablation}
\end{table}

\section{Conclusion and Discussion}

In this paper, we introduce ProBias, a Directed Bipartite Graph Encoder that addresses the long-tail problem in ICD coding by injecting a probability-based bias derived from co-occurrence statistics into the attention module to enrich rare code representations. To ensure high-quality input for the graph encoder, we utilize a large language model to generate comprehensive code descriptions. Experiments on three benchmark datasets demonstrate ProBias achieves state-of-the-art performance.

With regard to future work, while our core co-occurrence encoding component relies on discretizing continuous probability values through binning, we could explore a learnable mapping function that operates directly on these continuous values. For instance, a small multi-layer perceptron (MLP) could be used to transform probability values into a bias term, which will enable more precise capture of the subtle variations in comorbidity strength. Besides, although LLM-generated code descriptions have enhanced long-tail performance, a promising direction is to integrate a gating mechanism to dynamically fuse these LLM-derived descriptions with original code names or ontological concepts (e.g., from UMLS). This approach would combine the descriptive richness of LLMs with the precision of formalized data, thereby further boosting the performance of this component.

\bibliographystyle{elsarticle-num-names} 

\bibliography{reference}

@inproceedings{zhou2020hierarchy,
  title={Hierarchy-aware global model for hierarchical text classification},
  author={Zhou, Jie and Ma, Chunping and Long, Dingkun and Xu, Guangwei and Ding, Ning and Zhang, Haoyu and Xie, Pengjun and Liu, Gongshen},
  booktitle={Proceedings of the 58th annual meeting of the association for computational linguistics},
  pages={1106--1117},
  year={2020}
}

@article{valderas2009defining,
	author = {Valderas, Jose M. and Starfield, Barbara and Sibbald, Bonnie and Salisbury, Chris and Roland, Martin},
	title = {Defining Comorbidity: Implications for Understanding Health and Health Services},
	volume = {7},
	number = {4},
	pages = {357--363},
	year = {2009},
	publisher = {The Annals of Family Medicine},
	journal = {The Annals of Family Medicine}
}

@article{wang2017learning,
  title={Learning to model the tail},
  author={Wang, Yu-Xiong and Ramanan, Deva and Hebert, Martial},
  journal={Advances in neural information processing systems},
  volume={30},
  year={2017}
}

@inproceedings{yang2023multi,
  title={Multi-label few-shot icd coding as autoregressive generation with prompt},
  author={Yang, Zhichao and Kwon, Sunjae and Yao, Zonghai and Yu, Hong},
  booktitle={Proceedings of the AAAI Conference on Artificial Intelligence},
  volume={37},
  pages={5366--5374},
  year={2023}
}

@inproceedings{larkey1996combining,
  title={Combining classifiers in text categorization},
  author={Larkey, Leah S and Croft, W Bruce},
  booktitle={Proceedings of the 19th annual international ACM SIGIR conference on Research and development in information retrieval},
  pages={289--297},
  year={1996}
}

@article{wang2024value,
  title={On the value of head labels in multi-label text classification},
  author={Wang, Haobo and Peng, Cheng and Dong, Hede and Feng, Lei and Liu, Weiwei and Hu, Tianlei and Chen, Ke and Chen, Gang},
  journal={ACM Transactions on Knowledge Discovery from Data},
  volume={18},
  number={5},
  pages={1--21},
  year={2024},
  publisher={ACM New York, NY}
}

@inproceedings{xiao2021does,
  title={Does head label help for long-tailed multi-label text classification},
  author={Xiao, Lin and Zhang, Xiangliang and Jing, Liping and Huang, Chi and Song, Mingyang},
  booktitle={Proceedings of the AAAI conference on artificial intelligence},
  volume={35},
  pages={14103--14111},
  year={2021}
}

@inproceedings{liu2019large,
  title={Large-scale long-tailed recognition in an open world},
  author={Liu, Ziwei and Miao, Zhongqi and Zhan, Xiaohang and Wang, Jiayun and Gong, Boqing and Yu, Stella X},
  booktitle={Proceedings of the IEEE/CVF conference on computer vision and pattern recognition},
  pages={2537--2546},
  year={2019}
}

@article{raffel2020exploring,
  title={Exploring the limits of transfer learning with a unified text-to-text transformer},
  author={Raffel, Colin and Shazeer, Noam and Roberts, Adam and Lee, Katherine and Narang, Sharan and Matena, Michael and Zhou, Yanqi and Li, Wei and Liu, Peter J},
  journal={Journal of machine learning research},
  volume={21},
  number={140},
  pages={1--67},
  year={2020}
}

@inproceedings{shaw2018self,
  title={Self-Attention with Relative Position Representations},
  author={Shaw, Peter and Uszkoreit, Jakob and Vaswani, Ashish},
  booktitle={Proceedings of the 2018 Conference of the North American Chapter of the Association for Computational Linguistics: Human Language Technologies, Volume 2 (Short Papers)},
  pages={464--468},
  year={2018}
}

@inproceedings{xie2019ehr,
  title={EHR coding with multi-scale feature attention and structured knowledge graph propagation},
  author={Xie, Xiancheng and Xiong, Yun and Yu, Philip S and Zhu, Yangyong},
  booktitle={Proceedings of the 28th ACM international conference on information and knowledge management},
  pages={649--658},
  year={2019}
}

@inproceedings{gomes2024accurate,
  title={Accurate and Well-Calibrated ICD Code Assignment Through Attention Over Diverse Label Embeddings},
  author={Gomes, Goncalo and Coutinho, Isabel and Martins, Bruno},
  booktitle={Proceedings of the 18th Conference of the European Chapter of the Association for Computational Linguistics (Volume 1: Long Papers)},
  pages={2302--2315},
  year={2024}
}

@misc{johnson2016mimiciii,
  author = {Johnson, Alistair and Pollard, Tom and Mark, Roger},
  title = {{MIMIC-III Clinical Database} (version 1.4)},
  year = {2016},
  howpublished = {\url{https://doi.org/10.13026/C2XW26}},
  note = {Accessed: 2024-08-28}
}

@inproceedings{vaswani2017attention,
  title={Attention is all you need},
  author={Vaswani, Ashish and Shazeer, Noam and Parmar, Niki and Uszkoreit, Jakob and Jones, Llion and Gomez, Aidan N and Kaiser, Lukasz and Polosukhin, Illia},
  booktitle={Proc. of the 31st Int. Conference on Neural Information Processing Systems},
  pages={6000--6010},
  year={2017}
}

@article{ying2021transformers,
  title={Do transformers really perform badly for graph representation?},
  author={Ying, Chengxuan and Cai, Tianle and Luo, Shengjie and Zheng, Shuxin and Ke, Guolin and He, Di and Shen, Yanming and Liu, Tie-Yan},
  journal={Advances in neural information processing systems},
  volume={34},
  pages={28877--28888},
  year={2021}
}

@inproceedings{kipf2017semi,
  title={Semi-Supervised Classification with Graph Convolutional Networks},
  author={Kipf, Thomas N and Welling, Max},
  booktitle={International Conference on Learning Representations},
  year={2017}
}

@article{dwivedi2020generalization,
  title={A generalization of transformer networks to graphs},
  author={Dwivedi, Vijay Prakash and Bresson, Xavier},
  journal={arXiv preprint arXiv:2012.09699},
  year={2020}
}

@article{yang2022gatortron,
  title={Gatortron: A large clinical language model to unlock patient information from unstructured electronic health records},
  author={Yang, Xi and Chen, Aokun and PourNejatian, Nima and Shin, Hoo Chang and Smith, Kaleb E and Parisien, Christopher and Compas, Colin and Martin, Cheryl and Flores, Mona G and Zhang, Ying and others},
  journal={arXiv preprint arXiv:2203.03540},
  year={2022}
}

@article{liu2024lost,
  title={Lost in the middle: How language models use long contexts},
  author={Liu, Nelson F and Lin, Kevin and Hewitt, John and Paranjape, Ashwin and Bevilacqua, Michele and Petroni, Fabio and Liang, Percy},
  journal={Transactions of the Association for Computational Linguistics},
  volume={12},
  pages={157--173},
  year={2024},
  publisher={MIT Press One Broadway, 12th Floor, Cambridge, Massachusetts 02142, USA~…}
}

@inproceedings{mullenbach2018explainable,
  title={Explainable Prediction of Medical Codes from Clinical Text},
  author={Mullenbach, James and Wiegreffe, Sarah and Duke, Jon and Sun, Jimeng and Eisenstein, Jacob},
  booktitle={Proceedings of the 2018 Conference of the North American Chapter of the Association for Computational Linguistics: Human Language Technologies, Volume 1 (Long Papers)},
  pages={1101--1111},
  year={2018}
}

@inproceedings{cao2020hypercore,
  title={HyperCore: Hyperbolic and co-graph representation for automatic ICD coding},
  author={Cao, Pengfei and Chen, Yubo and Liu, Kang and Zhao, Jun and Liu, Shengping and Chong, Weifeng},
  booktitle={Proceedings of the 58th Annual Meeting of the Association for Computational Linguistics},
  pages={3105--3114},
  year={2020}
}

@inproceedings{vu2021label,
  title={A label attention model for ICD coding from clinical text},
  author={Vu, Thanh and Nguyen, Dat Quoc and Nguyen, Anthony},
  booktitle={Proceedings of the Twenty-Ninth International Conference on International Joint Conferences on Artificial Intelligence},
  pages={3335--3341},
  year={2021}
}

@inproceedings{yuan2022code,
  title={Code Synonyms Do Matter: Multiple Synonyms Matching Network for Automatic ICD Coding},
  author={Yuan, Zheng and Tan, Chuanqi and Huang, Songfang},
  booktitle={Proceedings of the 60th Annual Meeting of the Association for Computational Linguistics (Volume 2: Short Papers)},
  pages={808--814},
  year={2022}
}

@article{nguyen2023mimic,
  title={Mimic-iv-icd: A new benchmark for extreme multilabel classification},
  author={Nguyen, Thanh-Tung and Schlegel, Viktor and Kashyap, Abhinav and Winkler, Stefan and Huang, Shao-Syuan and Liu, Jie-Jyun and Lin, Chih-Jen},
  journal={arXiv preprint arXiv:2304.13998},
  year={2023}
}

@inproceedings{yang2022knowledge,
  title={Knowledge injected prompt based fine-tuning for multi-label few-shot icd coding},
  author={Yang, Zhichao and Wang, Shufan and Rawat, Bhanu Pratap Singh and Mitra, Avijit and Yu, Hong},
  booktitle={Proceedings of the conference on empirical methods in natural language processing. Conference on empirical methods in natural language processing},
  volume={2022},
  pages={1767},
  year={2022},
  organization={NIH Public Access}
}

@inproceedings{huang2022plm,
  title={PLM-ICD: Automatic ICD Coding with Pretrained Language Models},
  author={Huang, Chao-Wei and Tsai, Shang-Chi and Chen, Yun-Nung},
  booktitle={Proceedings of the 4th Clinical Natural Language Processing Workshop},
  pages={10--20},
  year={2022}
}

@inproceedings{luo2024corelation,
  title={CoRelation: Boosting Automatic ICD Coding through Contextualized Code Relation Learning},
  author={Luo, Junyu and Wang, Xiaochen and Wang, Jiaqi and Chang, Aofei and Wang, Yaqing and Ma, Fenglong},
  booktitle={Proceedings of the 2024 Joint International Conference on Computational Linguistics, Language Resources and Evaluation (LREC-COLING 2024)},
  pages={3997--4007},
  year={2024}
}

@inproceedings{nguyen2023two,
  title={A Two-Stage Decoder for Efficient ICD Coding},
  author={Nguyen, Thanh-Tung and Schlegel, Viktor and Kashyap, Abhinav Ramesh and Winkler, Stefan},
  booktitle={Findings of the Association for Computational Linguistics: ACL 2023},
  pages={4658--4665},
  year={2023}
}

@inproceedings{wang2024multi,
  title={Multi-stage Retrieve and Re-rank Model for Automatic Medical Coding Recommendation},
  author={Wang, Xindi and Mercer, Robert and Rudzicz, Frank},
  booktitle={Proceedings of the 2024 Conference of the North American Chapter of the Association for Computational Linguistics: Human Language Technologies (Volume 1: Long Papers)},
  pages={4881--4891},
  year={2024}
}

@inproceedings{rios2018few,
  title={Few-shot and zero-shot multi-label learning for structured label spaces},
  author={Rios, Anthony and Kavuluru, Ramakanth},
  booktitle={Proceedings of the conference on empirical methods in natural language processing. Conference on empirical methods in natural language processing},
  volume={2018},
  pages={3132},
  year={2018}
}

@misc{johnson2023mimiciv_note,
  author = {Johnson, Alistair and Pollard, Tom and Horng, Stephen and Celi, Leo Anthony and Mark, Roger},
  title = {{MIMIC-IV-Note: Deidentified free-text clinical notes} (version 2.2)},
  year = {2023},
  howpublished = {\url{https://doi.org/10.13026/1n74-ne17}},
  note = {Accessed: 2024-08-28}
}

\end{document}